\documentclass[acmsmall,screen,nonacm]{acmart}

\AtBeginDocument{%
  }

\usepackage[T1]{fontenc}
\usepackage[frozencache,cachedir=.]{minted}
\usepackage{tikz}
\usetikzlibrary{automata,arrows,positioning}
\usepackage{multirow}
\usepackage{subfigure}
\usepackage{makecell}
\usepackage{algorithmic}
\usepackage{xcolor,xspace}
\usepackage{mathtools}
\usepackage{colortbl}
\usepackage{listings}
\usepackage{enumitem}
\usepackage{hhline}
\usepackage{textcomp}
\usepackage{xcolor}
\usepackage{booktabs}
\usepackage{hyperref}
\usepackage{graphicx}
\usepackage{relsize}
\usepackage{makecell}
\usepackage{wrapfig}
\usepackage{bbding}
\usepackage{algorithmic}
\usepackage[algoruled,noend,linesnumbered]{algorithm2e}%

\newcommand{\nm}{\m{nm}}

\newcommand{\Obj}{o} 
\newcommand{\Subj}{s}
\newcommand{\m}{\mathit}

\usepackage{fancyvrb}
\usepackage{framed}

\definecolor{shadecolor}{rgb}{0.85,0.85,0.85}
\definecolor{backgroundcolor}{rgb}{0.95,0.95,0.92}
\definecolor{keywordcolor}{rgb}{0.13,0.29,0.53}
\definecolor{errorbackground}{rgb}{1.0,0.7,0.7}
\definecolor{warningbackground}{rgb}{1.0,0.84,0.5}
\definecolor{commentcolor}{rgb}{0.72, 0.25, 0.05}
\definecolor{BrickRed}{RGB}{170,74,68}
\definecolor{ForestGreen}{RGB}{34,139,34}
\definecolor{Maroon}{RGB}{128,0,0}

\newcommand{\error}[1]{\colorbox{errorbackground}{#1}}

\lstdefinelanguage{Isar}%
{morekeywords={
_first,abbreviation,then,theory,begin,infix,where,for,fun,apply,assumes,done,lemma,shows,sorry,proof,unfolding,using,by,let,have,and,fix,thus,next,assume,imports,fixes,also,finally,show,sledgehammer,qed,_last
    },
sensitive=true,
morecomment=[s]{(*}{*)},
literate=
    {\\<not>}{{$\neg$}}1
{\\<notin>}{{$\not\in$}}1
{\\<times>}{{$×$}}1
{\\<Rightarrow>}{{$\Rightarrow$}}1
{\\<equiv>}{{$\equiv$}}1
{~=}{{$\not=$}}1
{\\<rightleftharpoons>}
{{$\rightleftharpoons$}}1
{\\<exists>}{{$\exists$}}1
{\\<parallel>}{{$\parallel$}}1
{\\<Longrightarrow>}
{{$\Longrightarrow$}}1
{\\<lbrakk>}{{$\lsemantics$}}1
{\\<rbrakk>}{{$\rsemantics$}}1
{\\<longrightarrow>}
{{$\longrightarrow$}}1
{\\<and>}{{$\land$}}1
{\\<or>}{{$\lor$}}1
{\\<bigwedge>}{{$\bigwedge$}}1
{\\<le>}{{$\le$}}1
{\\<mapsto>}{{$\mapsto$}}1
{\\<in>}{{$\in$}}1
{\\<approx>}{{$\approx$}}1
{\\<subseteq>}{{$\subseteq$}}1
{\\<pi>}{{$\pi$}}1
{\\<lambda>}{{$\lambda$}}1
{\\<Theta>}{{$\Theta$}}1
{\\<land>}{{$\land$}}1
{\\<pal>}{{$\models_!$}}1
{\\<palbot>}{{$\bot_!$}}1
{\\<union>}{{$\cup$}}1
{\\<And>}{{$\bigwedge$}}1
{!!}{{$\bigwedge$}}1
{ü}{"u}1 {ä}{"a}1 {ö}{"o}1
{Ü}{"U}1 {Ä}{"A}1 {Ö}{"O}1 {ß}{{ss}}1
}[keywords,comments,strings]%

\lstdefinelanguage{Coq}%
{morekeywords={
_first,Theorem,Proof,Qed,intros,unfold,destruct,as,exists,rewrite,ring,in,_last
    },
sensitive=true,
morecomment=[s]{(*}{*)}
}[keywords,comments,strings]%

\newenvironment{codeframe}{%
    \setlength{\fboxsep}{0pt}%
    \setlength{\fboxrule}{0pt}%
    \MakeFramed {\FrameRestore}%
}{
    \endMakeFramed
}
\begin{document}

\title{The Fusion of Large Language Models and Formal Methods for Trustworthy AI Agents: A Roadmap}

\author{Yedi Zhang$^{\dagger\ast}$}
\email{yd.zhang@nus.edu.sg}

\author{Yufan Cai$^\ast$}
\author{Xinyue Zuo$^\ast$}
\affiliation{%
  \institution{National University of Singapore}
  \country{Singapore}
}

\author{Xiaokun Luan$^\ast$}
\affiliation{%
  \institution{Peking University}
  \country{China}
}

\author{Kailong Wang$^\ast$}
\email{wangkl@hust.edu.cn}
\affiliation{%
  \institution{Huazhong University of Science and Technology}
  \country{China}
}

\author{Zh\'e H\'ou$^\ast$}
\email{z.hou@griffith.edu.au}
\affiliation{%
  \institution{Griffith University}
  \country{Australia}
}

\author{Yifan Zhang}
\affiliation{%
  \institution{National University of Singapore}
  \country{Singapore}
}

\author{Zhiyuan Wei}
\affiliation{%
  \institution{Beijing Institute of Technology}
  \country{China}
}

\author{Meng Sun}
\affiliation{%
  \institution{Peking University}
  \country{China}
}

\author{Jun Sun}
\affiliation{%
  \institution{Singapore Management University}
  \country{Singapore}
}

\author{Jing Sun}
\affiliation{%
  \institution{University of Auckland}
  \country{New Zealand}
}

\author{Jin Song Dong}
\email{dcsdjs@nus.edu.sg}
\affiliation{%
  \institution{National University of Singapore}
  \country{Singapore}
}

\renewcommand{\shortauthors}{Zhang et al.}


\begin{abstract}
Large Language Models (LLMs) have emerged as a transformative AI paradigm, profoundly influencing daily life through their exceptional language understanding and contextual generation capabilities. 
Despite their remarkable performance, LLMs face a critical challenge: the propensity to produce unreliable outputs due to the inherent limitations of their learning-based nature. Formal methods (FMs), on the other hand, are a well-established computation paradigm that provides mathematically rigorous techniques for modeling, specifying, and verifying the correctness of systems.
FMs have been extensively applied in mission-critical software engineering, embedded systems, and cybersecurity. 
However, the primary challenge impeding the deployment of FMs in real-world settings lies in their steep learning curves, the absence of user-friendly interfaces, and issues with efficiency and adaptability. 

This position paper outlines a roadmap for advancing the next generation of trustworthy AI systems by leveraging the mutual enhancement of LLMs and FMs. 
First, we illustrate how FMs, including reasoning and certification techniques, can help LLMs generate more reliable and formally certified outputs. 
Subsequently, we highlight how the advanced learning capabilities and adaptability of LLMs can significantly enhance the usability, efficiency, and scalability of existing FM tools. 
Finally, we show that unifying these two computation paradigms---integrating the flexibility and intelligence of LLMs with the rigorous reasoning abilities of FMs---has transformative potential for the development of trustworthy AI software systems. We acknowledge that this integration has the potential to enhance both the trustworthiness and efficiency of software engineering practices while fostering the development of intelligent FM tools capable of addressing complex yet real-world challenges.


\end{abstract}

%
%
    
    
   



\settopmatter{printfolios=true,printccs=false,printacmref=false}
\renewcommand\footnotetextcopyrightpermission[1]{}

%

\thanks{$^\ast$Equal contribution, $^\dagger$Corresponding author}

\maketitle

\section{Introduction}\label{sec:intro}
The rapid advancement of modern AI techniques, particularly in the realm of Large Language Models (LLMs) like GPT~\cite{achiam2023gpt}, Llama~\cite{touvron2023llama}, Claude~\cite{TheC3}, Gemini~\cite{geminiteam2024gemini}, etc., has marked a significant evolution in human-level computational capabilities. These models are fundamentally reshaping tasks across a spectrum of applications, from natural language processing to automated content generation. Trained on vast text corpora, LLMs excel in generating responses that are contextually accurate and stylistically appropriate. However, their applicability in safety-critical or knowledge-critical settings remains limited due to their inherent reliability issues---primarily, their propensity for generating outputs that, while plausible, may be factually incorrect or contextually inappropriate~\cite{jacovi2020towards,wiegreffe1teach,agarwal2024faithfulness}. This limitation, known as ``hallucination'', stems from the probabilistic nature of learning-based AI, where the models optimize for likelihood rather than truth or logical consistency. 
%

In a completely different thread of research, Formal Methods (FMs) have been established as rigorous tools for verification and validation (V\&V) of critical systems where failure is intolerable, such as aerospace (relevant areas including avionics)~\cite{dragomir2022model,liu2019aadl}, autonomous driving~\cite{konig2024towards,alves2021double,huang2022model}, and medical devices~\cite{jetley2006formal,freitas2020medicine,arcaini2018integrating}. By performing appropriate mathematical analysis, these methods are designed to ensure the correctness and safety of both hardware and/or software systems. Despite their demonstrated benefits, the adoption of FMs remains limited, primarily due to their significant computational complexity and the specialized expertise required for their effective and efficient implementation.

Although both computational paradigms encounter inherent challenges of their own---namely, the unreliability stemming from the statistical nature of LLMs and the user-unfriendly interfaces coupled with the computational complexity of formal methods---recent studies have highlighted their potential for mutual benefits~\cite{wu2022autoformalization,pan2023logic,he2023solving,ZhouSLSWW24,yang2019learning,yang2024leandojo,song2024towards}. Efforts to bridge these two paradigms aim to harness their respective strengths, with the ultimate objective of developing a neural-symbolic AI seamlessly integrates LLMs and FMs into a unified solution. For instance, to enhance the reliability of LLMs, various approaches~\cite{pan2023logic,ma2024sciagent} have incorporated solvers to facilitate reasoning tasks guided by specification rules or reasoning models derived from LLMs' inputs/outputs. Specifically, some efforts have been directed toward improving LLMs' understanding of Lisp, enabling better integration with Lisp-based programming techniques~\cite{stengeleskin2024,li2024guiding}. Conversely, within the formal methods community, there is a growing trend to leverage LLMs to enhance the functionality and usability of automated verification approaches~\cite{wuLemur,wen2024enchanting}. Indeed, LLMs are increasingly utilized to automate and simplify the generation of specifications and proofs, and by employing LLMs to analyze natural language requirements or extract logical properties from complex systems, researchers aim to bridge the gap between human-readable specifications and machine-executable formal models.

In this paper, we aim to advance the research agenda of bi-directional enhancement by building on insights from our prior investigations and providing a comprehensive discussion of its implications, technical challenges, and potential future directions. 
More specifically, we first look into how FMs can contribute to improving the trustworthiness of LLMs.
In this context, we discuss how symbolic solvers, such as SMT solvers, can be seamlessly integrated into LLMs to enable a complete reasoning cycle, thereby enhancing their logical reasoning capabilities and ensuring more robust and reliable responses. Additionally, we investigate methods for advancing the evaluation of LLMs through logical reasoning frameworks and delve into rigorous behavior analysis of LLMs via formal verification techniques. 
Next, we examine how LLMs can drive advancements in existing formal methods and tools, focusing on three key areas: (i) utilizing LLMs for the automated formalization of specifications to bridge natural language and formal representations, (ii) leveraging LLMs to enhance theorem proving by automating proof generation and validation, and (iii) employing LLMs to develop intelligent model checkers with improved adaptability and efficiency. 
Finally, we recognize that the bi-directional interactions between LLMs and FMs open avenues for further exploration, highlighting their potential to drive the development of next-generation AI systems that seamlessly integrate these paradigms. We believe that such integration holds significant promise for developing trustworthy AI agents across diverse application domains, as exemplified by the widely applicable scenario of program synthesis.

The remainder of this paper is structured as follows: Section~\ref{sec:roadmap} outlines a roadmap for building trustworthy agents by combining the strengths of FMs and LLMs. Sections~\ref{sec:fm4llm} and~\ref{sec:llm4fm} explore their bi-directional benefits, focusing on FMs for LLMs and LLMs for FMs, respectively. Section~\ref{sec:fm_llm} highlights our exploration of integrating LLMs and FMs as a unified computational paradigm. Finally, Section~\ref{sec:con} concludes the paper.

\section{A Roadmap to Trustworthy LLM: Combining LLM and FM}\label{sec:roadmap}

This section gives a roadmap for the development of a trustworthy LLM agent, specifically tailored to tasks involving FMs, emphasizing the integration of rigorous formal analysis techniques with the adaptability and learning capabilities of language models.

\begin{figure}[t]
    \centering
    \includegraphics[width=0.6\linewidth]{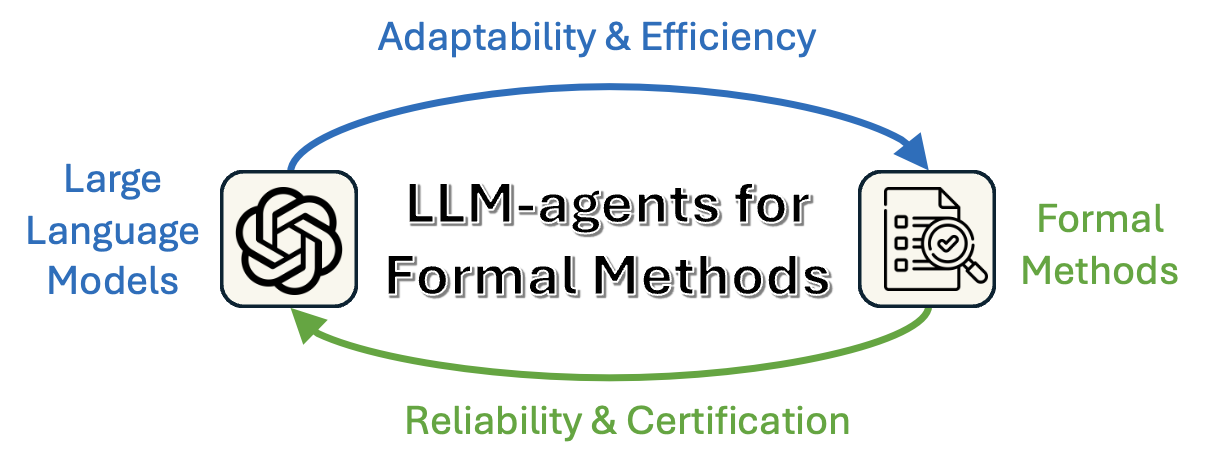}
    \caption{Towards Trustworthy LLM Agents with Formal Methods.}
    \label{fig:roadmap}
\end{figure}

The proposed roadmap, depicted in Figure~\ref{fig:roadmap}, adopts a twofold strategy. 
On the one hand,  we advocate for the integration of formal analysis principles throughout the development and V\&V phases. This aims to enhance the reasoning capabilities of LLM agents while ensuring their reliability and compliance with stringent certification standards, a strategy direction we term \emph{FMs for LLMs}.
On the other hand, we leverage the adaptability and learning capabilities of LLMs during the functionality development phase of FM-based tools. This is aimed at enhancing the flexibility and reasoning efficiency of established formal analysis methodologies, such as theorem proving and model checking. In this paper, we refer to this strategic direction as \emph{LLMs for FMs}. Ultimately, we argue that the dynamic interplay between these two strategic directions will substantially advance the development of trustworthy LLM agents, particularly in the domain of formal methods. 

Building on these foundations, we outline the envisioned characteristics of a trustworthy LLM agent as follows.
Firstly, an LLM agent tailored for formal analysis should be implemented directly at the client's premises, allowing them to learn from and adapt to the client's specific data and workflows, also allowing the client's domain-specific experts to use formal methods through LLM agents with a low entry bar. This localized learning approach should not only enable the agents to enhance their performance over time but also ensure that the entire process safeguards the client's privacy. Based on this, LLM agents will significantly reduce both the human learning curve and the requisite human resources needed to apply formal methods, thereby facilitating a more efficient integration of these methodologies into the client's operational framework. 
Secondly, each output suggested by an LLM agent should be both faithful and rigorously certified, particularly when the response is expected to be deterministic. This requirement ensures that the outputs not only align closely with the rules defined within the formal methods framework but also meet predefined accuracy and reliability standards. To achieve this, the outputs must undergo a thorough certification process, possibly involving techniques such as testing or verification, to validate that they are correct and adhere strictly to the operational and regulatory requirements. Note that, such a certification process is critical, as it guarantees that the solutions provided by the LLM agents are not only effective but also trustworthy and legally compliant, thereby increasing the confidence of stakeholders in deploying these agents within sensitive or critical operational contexts. 

In the subsequent sections, we will adhere to the roadmap outlined in Figure~\ref{fig:roadmap} to present our insights, document our work in progress, and outline our future visions regarding the two strategic directions. Additionally, we will also discuss the synergistic outcomes derived from the integration of these two computation paradigms. Specifically, we demonstrate how this integration can lead to the development of programming refinement methodologies, exemplifying their applications in more intelligent computational environments.



\section{FM for LLM: Towards Reliability and Certification}\label{sec:fm4llm}

In this section, we illustrate how formal methods can enhance the reliability and help certification of LLMs. 
Specifically, we explore this integration direction from three perspectives: (i) trustworthy LLMs with symbolic solvers, (ii) LLM Testing based on logical reasoning, and (iii) rigorous LLM behavior analysis. 
We argue that incorporating these FM-based techniques enables AI systems to become reliably secure, paving the way for developing trustworthy AI systems.

Note that this section highlights the initial efforts to apply formal methods to enhance the reliability of LLMs, acknowledging that these approaches are not exhaustive. While a fully comprehensive methodology for integrating formal methods into LLMs remains elusive, we present our current implementations and future visions as preliminary steps toward advancing this direction.

\subsection{SMT Solvers for LLMs}
Satisfiability Modulo Theories (SMT) solvers are specialized tools designed to determine the satisfiability of logical formulas defined over some theories, such as arithmetic, bit-vectors, and arrays. 
They play a pivotal role in formal verification, program analysis, and automated reasoning, serving as essential components to ensure the correctness and reliability of complex software systems.

Recent studies~\cite{deng2024enhancing,pan2023logic,wang2024dataflow,ye2024satlm} have explored the integration of SMT solvers to enhance the accuracy and reliability of LLMs in logic reasoning tasks. These solver-powered LLM agents operate by translating task descriptions into formal specifications, delegating reasoning tasks to specialized expert tools for precise analysis, and subsequently converting the outputs back into natural language. 

We have identified three main challenges within this research line. Firstly, while LLMs are capable of generating logical constraints or SMT formulas, they often produce suboptimal or overly verbose constraints, which can place an additional computational burden on the solver. 
Secondly, the outputs of LLMs lack guarantees of correctness or logical consistency, potentially introducing subtle inaccuracies or ambiguities in the generated SMT constraints. It can lead to invalid results or solutions that are challenging to interpret.
Lastly, LLMs often lack domain-specific knowledge and may struggle to generate outputs that conform to the precise formal syntax required by SMT solvers. Consequently, they may generate formulas that are semantically sensible but syntactically invalid formulas, rendering them unprocessable by the solvers.

\subsubsection{Our Insights}
We present our insights and proposed strategies to address the three key challenges outlined above.

\smallskip
\noindent
{\bf Strategy 1. Multiple LLMs Debating.} To address the challenge of LLMs generating suboptimal or overly verbose constraints, a potential strategy involves leveraging multiple LLMs in a collaborative or adversarial framework to critique, validate, and refine each other's outputs.
In this approach, the system employs one or more LLMs to generate SMT code from natural language inputs, while other LLMs function as ``critics'', evaluating the generated SMT code for logical consistency, syntactic correctness, and alignment with the problem description. By incorporating feedback loops among these models, the system can iteratively refine the outputs and reduce ambiguity inherent in natural language inputs. 

\noindent
{\bf  Strategy 2. Test Generation.} 
Test cases will be automatically generated to validate the correctness and consistency of the LLM-generated SMT code against the expected behavior. Fuzzing techniques may also be employed to generate adversarial inputs for testing. 
Additionally, mutation-based approaches can be applied to both the SMT code and the natural language descriptions, with two LLMs comparing the resulting solutions. 
The strategy helps check the consistency between the natural language description and the SMT code produced by the LLMs.

\noindent
{\bf  Strategy 3. Self-Correction.} 
Feedback from tests, critics, or the solver itself can be leveraged to iteratively refine the SMT code. 
Errors identified via solvers can be categorized into syntax issues (e.g., invalid SMT-LIB syntax), semantic misalignments (e.g., logical inconsistencies), or performance bottlenecks (e.g., slow or incomplete solver responses). 
Based on this feedback, an LLM can be employed to debug and regenerate problematic parts of the constraints, ensuring that the refinements are both targeted and context-aware. 
This iterative refinement process, coupled with validation through re-testing, facilitates the convergence of LLM-generated SMT codes toward correctness and rigorousness.

\subsubsection{Case Study on LLMs with Z3}
To illustrate our perspective, we give our recent exploration of the interaction between LLMs and Z3 in Python. 

Z3~\cite{z3}, a widely used SMT solver, accepts inputs in the form of simple-sorted formulas expressed in first-order logic (FOL). These formulas can include symbols with predefined meanings, defined by the underlying theories supported by the solver, and these theories encompass domains such as arithmetic, bit-vectors, arrays, etc., making Z3 particularly well-suited for reasoning about a wide range of logical constraints. 

\paragraph{Example} Consider a scenario where a user requests the LLM agent to solve a scheduling problem: 

\begin{quote}
\textit{``Can you help plan a meeting for a team of three people—David, Emma, and Alex? David is free on Monday or Tuesday, Emma is free on Tuesday or Wednesday, and Alex is free only on Tuesday or Thursday. Find a common day when all three are available.''}
\end{quote}

We now provide a detailed, step-by-step solution for this task:

\paragraph{Formalization of Constraints} Given the above problem, the initial Z3 constraints in Python generated by the LLM are given as follows:
\begin{minted}[fontsize=\footnotesize,bgcolor=backgroundcolor]{python}
    # Define days of the week
    days = ["Monday", "Tuesday", "Wednesday", "Thursday"]
    David_free = [Bool(f"David_free_{day}") for day in days]
    Emma_free = [Bool(f"Emma_free_{day}") for day in days]
    Alex_free = [Bool(f"Alex_free_{day}") for day in days]
    # Create a solver
    solver = Solver()
    # Define constraints for each person's availability
    solver.add(Or(David_free[0], David_free[1])) 
    solver.add(Or(Emma_free[1], Emma_free[2]))  
    solver.add(Or(Alex_free[1], Alex_free[3]))   
    # Add constraints that ensure a common day for all three
    common_day_constraints = [And(David_free[i], 
    Emma_free[i], Alex_free[i]) for i in range(len(days))]
    solver.add(Or(common_day_constraints)) 
\end{minted}

\paragraph{Self correction}
If the Z3 code has issues (e.g., missing constraints or syntax errors) or generates inconsistent results with the natural language description, the self-correction procedure will identify and correct them.
In this example, the previous Z3 code ignores the following constraints:
\begin{minted}[fontsize=\footnotesize,bgcolor=backgroundcolor]{python}
    # Constraints for David
    solver.add(And(Not(David_free[2]), Not(David_free[3]))) 
    # Constraints for Emma
    solver.add(And(Not(Emma_free[0]), Not(Emma_free[3])))  
    # Constraints for Alex
    solver.add(And(Not(Alex_free[0]), Not(Alex_free[2])))  
\end{minted}

\paragraph{Test Generation}
The agent mutates the constraints and tweaks the availability of each individual to create new conditions.
For example, the new mutated constraints are
David will be free on Monday and Wednesday.
Emma will be free on Tuesday and Thursday.
Alex will be free on Monday and Thursday.
The updated Z3 code generated by the LLM is as follows:
\begin{minted}[fontsize=\footnotesize,bgcolor=backgroundcolor]{python}
    # Mutated constraints for David
    solver.add(And(David_free[0], David_free[2])) 
    solver.add(And(Not(David_free[1]), Not(David_free[3]))) 
    # Mutated constraints for Emma
    solver.add(And(Emma_free[1], Emma_free[3]))  
    solver.add(And(Not(Emma_free[0]), Not(Emma_free[2]))) 
    # Mutated constraints for Alex
    solver.add(And(Alex_free[0], Alex_free[3])) 
    solver.add(And(Not(Alex_free[1]), Not(Alex_free[2]))) 
\end{minted}
The agent systems will check the consistency between the results produced by Z3 and the reasoning derived from natural language descriptions to further ensure the correctness of the Z3 codes.

\paragraph{Multiple LLM Debating}
Whenever it comes to a collision between the Z3 reasoning results and the natural language reasoning results, the LLM debating will be activated to debate which part is correct.
For example, after LLM-A generates the initial constraints and gets the results of Z3 code.
LLM-B will critique the constraints, identifying potential issues such as missing exclusivity rules or improperly translated logic.
LLM-C can suggest refinements, such as introducing mutual exclusivity or expanding constraints to handle edge cases.
The consensus will be the output with the highest confidence score (e.g., most accurate or simplest) is selected for testing and execution. 

\paragraph{Problem Solving} The translated constraints are fed into the Z3 solver, which checks the satisfiability of the formula and computes a solution if possible.
\begin{minted}[fontsize=\footnotesize,bgcolor=backgroundcolor]{python}
    # Check for a solution
    if solver.check() == sat:
        model = solver.model()
        common_days = [days[i] for i in range(len(days)) 
        if model.evaluate(David_free[i]) 
           and model.evaluate(Emma_free[i]) 
           and model.evaluate(Alex_free[i])]
        print(f"Common day(s) when everyone is free:
        {common_days}")
    else:
        print("No common day when everyone is free.")
\end{minted}

\paragraph{Solution Interpretation} The LLM agent receives the solution from the Z3 solver and translates it back into natural language for the user.
The only day when all three are free is Tuesday. The output will be: Common day(s) when everyone is free: [`Tuesday'].

\subsection{Logical Reasoning for LLM on Test Case Generation}

LLM Testing~\cite{zhong2024agieval,HendrycksBBZMSS21,HuangBZZZSLLZLF23,zhou2023don} is primarily focused on establishing a comprehensive benchmark to evaluate the overall performance of the models, ensuring that they fulfill specific assessment criteria, such as accuracy, coherence, fairness, and safety, in alignment with their intended applications. Among these criteria, the phenomenon of ``hallucination'' is particularly noteworthy. This occurs when LLMs generate coherent but factually or contextually inaccurate or irrelevant output during tasks such as problem-solving. 

Recent research has introduced a variety of methodologies for detection and evaluation to mitigate hallucination issues in LLMs. A common and straightforward method is to create comprehensive benchmarks specifically designed to assess LLM performance. However, these methods, which often rely on simplistic or semi-automated techniques such as string matching, manual validation, or cross-verification using another LLM, has significant shortcomings in automatically and effectively testing Fact-conflicting hallucinations (FCH)~\cite{li2024hallu}. This is largely due to the lack of dedicated ground truth datasets and specific testing frameworks. We contend that unlike other types of hallucinations, which can be identified through checks for semantic consistency, FCH requires the verification of content's factual accuracy against external, authoritative knowledge sources or databases. Hence, it is crucial to automatically construct and update factual benchmarks, and automatically validate the LLM outputs based on that.

To this end, we propose to apply logical reasoning 
to design a reasoning-based test case generation method aimed at developing an extensive and extensible FCH testing framework. Such a testing framework leverages factual knowledge reasoning combined with metamorphic testing principles to ensure a comprehensive and robust FCH evaluation of LLM outputs.


\subsubsection{Overview}
A reasoning-based test case generation framework for FCH in LLMs should comprise the following four key modules working in sequential:





\smallskip
\noindent
{\bf Module 1. Factual Knowledge Extraction.} Utilizing extensive knowledge database dumps, the essential information and factual triples of verified entities are systematically extracted.

\noindent
{\bf Module 2. Logical Reasoning.} This module enables employing reasoning rules for generating valid and diverse facts and establishing new ground truth knowledge.

\noindent
{\bf Module 3. Benchmark Construction.} Focused on producing high-quality test case-oracle pairs from the newly generated ground truth knowledge, this module uses a straightforward metamorphic relation: \emph{Questions that align with the generated knowledge should receive a response of ``YES'', while questions that contradict it should be answered with ``NO''.} This module also includes methods for generating or selecting effective prompts to facilitate reliable interaction with LLMs.

\noindent
{\bf Module 4. Response Evaluation.} In the final module, the responses from LLMs are evaluated to detect factual consistency automatically. The module parses LLM outputs using NLP techniques to construct semantic-aware representations and assesses semantic similarity with ground truth. It then applies similarity-based oracles, using metamorphic testing to evaluate the consistency between LLM responses and established ground truth.






\subsubsection{Factual Knowledge Extraction}
This process focuses on extracting essential factual information from input knowledge data in the form of fact triples, which are then suitable for logical reasoning. Existing knowledge databases~\cite{freebase, DBpedia, Yago, WordNet}  serve as valuable resources due to their extensive repositories of structured data derived from documents and web pages. This structured data forms the foundation for constructing and enriching factual knowledge, providing a robust basis for the test case framework. The approach leverages the categorization of entities and relations defined in resources like Wikipedia, as demonstrated in accompanying figures that illustrate entity categories, relation types, and example fact triples.

The extraction process typically involves structuring facts as three-element predicates, $\nm\,(\Subj,\Obj)$, where ``$\Subj$'' (stands for $\m{subject}$) and ``$\Obj$'' (stands for $\m{object}$) are entities, and ``$\nm$''  denotes the predicate. This divide-and-conquer strategy extracts facts category by category, effectively organizing information across various domains. The extraction processs iterates through predefined categories of entities and relations, employing a database querying function to retrieve all relevant fact triples for a given entity and predicate combination. This ensures comprehensive and systematic extraction of factual knowledge, creating a well-structured dataset for reasoning and testing.

\subsubsection{Logical Reasoning}
This step focuses on deriving enriched information from previously extracted factual knowledge by employing logical reasoning techniques. The approach utilizes a logical programming-based processor to automatically generate new fact triples by applying predefined inference rules, taking one or more input triples and producing derived outputs.

In particular, to introduce variability in the generation of test cases, reasoning rules, commonly utilized in existing literature~\cite{zhou2019completing, liang2022reasoning, abboud2020boxe} for knowledge reasoning, are typically adopted, including negation, symmetric, inverse, transitive and composite. These rules provide a systematic framework for generating new factual knowledge, ensuring diverse and comprehensive test case preparation. The system applies all relevant reasoning rules exhaustively to the appropriate fact triples, enabling the automated enrichment of the knowledge base for further testing purposes.

\subsubsection{Benchmark Construction}
This process consists of two key steps: (i) generating question-answer (Q\&A) pairs and (ii) creating prompts from derived triples, which together can significantly reduce manual effort in test oracle generation.

\smallskip
\noindent
{\bf Step 1. Question Generation.} This step uses entity-relation mappings to populate predefined Q\&A templates, aligning relation types with corresponding question structures based on the grammatical and semantic characteristics of predicates. For predicates with unique characteristics, customized templates are employed to generate valid Q\&A pairs. To enhance natural language formulation, LLM can be used to refine the Q\&A structures. Answers are derived directly from factual triples, with true/false judgments determined by the data. Mutated templates, leveraging synonyms or antonyms, diversify questions with opposite semantics, yielding complementary answers.

\noindent
{\bf Step 2. Prompt Construction.} Prompts are designed to instruct LLMs to provide explicit judgments (e.g., yes/no/I don't know) and outline their reasoning in standardized formats. LLM analysts can utilize predefined instructions to ensure clarity and enable LLMs to deliver assessable and logically consistent responses. This method maximizes the model’s reasoning capabilities within the structured framework of prompts and cues.



\subsubsection{Response Evaluation.}
This step aims to enhance the factual evaluation in LLM outputs by identifying discrepancies between LLM responses and the verified ground truth in Q\&A pairs. Since the ``yes'' or ``no'' answers from LLMs may be unreliable, the module emphasizes analyzing the reasoning process to assess factual consistency. The approach uses semantic similarity between parsed LLM responses and ground truth to detect inconsistencies systematically.

The process begins with preliminary screening, where responses indicating the LLM's inability to answer (e.g., ``I don't know'') are classified as correct, as they align with the model's honesty principle. Suspicious responses proceed to response parsing and semantic structure construction, where statements in the reasoning process are parsed into triples. These triples are used to construct a semantic graph for the response, which is compared with a similar graph constructed from ground truth triples. 
Finally, similarity-based metamorphic testing and oracles evaluate consistency by comparing the semantic structures of the response and ground truth, focusing on node similarity (fact correctness) and edge similarity (reasoning correctness). Responses are categorized into four classes: correct responses (both nodes and edges are similar), hallucinations from erroneous inference (nodes are similar, edges are dissimilar), hallucinations from erroneous knowledge (edges are similar, nodes are dissimilar), and overlaps with both issues (both nodes and edges are dissimilar).



\subsection{Rigorous LLM Behavior Analysis}

While LLM Testing techniques can effectively provide broad assessments and reveal edge cases that may provoke unexpected responses, they are limited in their capability to give rigorous guarantees on LLM behaviors. LLM Verification, on the other hand, serves as a complementary mechanism. However, as LLMs grow more complex and tasks become increasingly sophisticated, traditional neural network verifiers lose relevance due to their limitations in accommodating diverse model architectures and their focus on single-application scenarios. Indeed, formal verification of LLMs poses intrinsic challenges due to three key factors: 

\smallskip
\noindent
{\bf Factor 1. Non-Deterministic Responses.}  Responses from LLMs are non-deterministic, meaning their outputs may vary even with the same input. This inherent variability presents substantial challenges to providing deterministic guarantees regarding their behavior.
    
\noindent
{\bf Factor 2. High Input Dimensions.} The high dimensionality of inputs in LLMs leads to exponential growth in the number of input tokens, rendering exhaustive verification across an infinite input space highly impractical.

\noindent
{\bf Factor 3. Lack of Formal Specification.} While formal specifications are rigorous, they often lack the expressive capability of natural language, which makes it extremely difficult to precisely capture the nuanced and complex language behaviors expected from LLMs.


\smallskip
Hence, we propose that a specialized verification paradigm tailored specifically for LLMs should be considered to ensure reliable and rigorous certification for long-term applications.

Given these challenges, we argue that monitoring might serve as a viable long-term solution for reliable LLM behavior analysis. Positioned between testing and verification, monitoring of formalized properties at runtime enables rigorous certification of system behavior with minimal computation overhead by examining execution traces against predefined properties. This approach has already inspired several efforts to monitor LLM responses at runtime~\cite{cohen2023,manakul2023selfcheckgpt,besta2024checkembed,chen2024complex} (quite similar to another research line named guardrails). However, the specifications used in these methods remain ambiguous and informal. For example, they define the properties of low hallucination based on the stability of LLM outputs. More recently, an approach~\cite{cheng2024formal} has been introduced to monitor the conditional fairness properties of LLM responses. The specifications in~\cite{cheng2024formal} are informed by linear temporal logic and its bounded metric interval temporal logic variant, reflecting a shift toward formal methods for more precise and dependable monitoring of LLM behavior. Despite these progresses, further efforts are needed to enable rigorous monitoring of a wider range of properties to fit in more real-world settings, which will remain our primary focus in the future.

\section{LLM for FM: Solving Verification Tasks Intelligently}\label{sec:llm4fm}

This section explores how LLMs can enhance formal methods by developing intelligent LLM agents for tasks such as theorem proving and model checking. 
These agents bring adaptability and efficiency to traditional formal verification processes, paving the way for more advanced and effective formal methods.

Formal methods are foundational techniques for ensuring system correctness and reliability, grounded in rigorous mathematical approaches such as theorem proving, model checking, and constraint solving. Despite their rigorous and provable guarantees, these techniques face significant barriers to industry adoption, primarily due to the complexity of formalizing requirement specifications, the limited scalability of algorithms for large systems, and the substantial manual effort involved in proof generation and validation.

Nevertheless, recent advances in LLMs present new opportunities to address these challenges. 
With their ability to process and generate natural language, structured code, and symbolic representations, LLMs can serve as intelligent assistants, automating tedious tasks and enhancing the capabilities of existing formal methods algorithms. This section delves into these possibilities through targeted case studies, demonstrating how LLMs can make formal methods more efficient and effective.

\subsection{LLM for Autoformalization}

Autoformalization is the process of automatically translating natural language based specifications or informal representations into formal specifications or proofs.
This complex task demands a deep understanding of both informal and formal languages, along with the ability to generate accurate, machine-readable formal representations.
The recent advances of LLMs have opened new avenues for automating the formalization process, which is traditionally a manual task.
Recent research has demonstrated the effectiveness of LLMs in various autoformalization scenarios, including neural theorem proving~\cite{jiang2022draft}, temporal logic generation~\cite{murphy2024guidingllmtemporallogic}, and program specification generation based on source code~\cite{ma2024specgen}.
In this section, we explore the role of LLM agents in facilitating proof autoformalization, focusing on the challenges and opportunities in this line of research.

Informal proofs, commonly found in textbooks, research papers, online forums, or even generated by LLMs, often omit details that human consider trivial or self-evident.
However, to ensure rigorous verification by theorem provers, they need to be translated into formal proofs that adhere to a specific syntax, where all the details are explicitly provided.
The challenge lies in bridging the gap between the informal proof sketches and the detailed rigor required in formal proofs.


\begin{figure}[t]
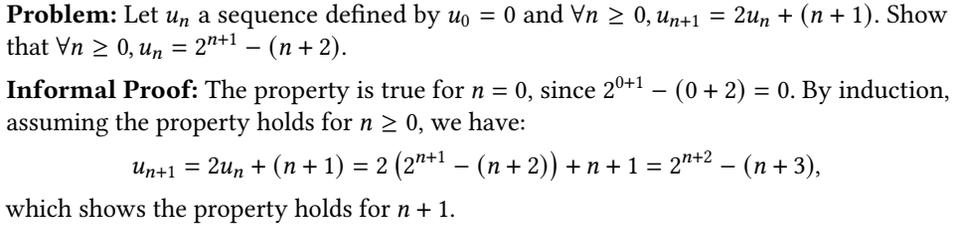

    \fbox{%
        \begin{minipage}[t]{0.9\linewidth}
            \textbf{Problem:} Let $u_n$ a sequence defined by $u_0 = 0$ and $\forall n \geq 0, u_{n+1} = 2 u_n + (n + 1)$. Show that $\forall n \geq 0, u_n = 2^{n+1} - (n+2)$.

            \medskip

            \textbf{Informal Proof:} The property is true for $n=0$, since $2^{0+1}-(0+2)=0$. By induction, assuming the property holds for $n\geq 0$, we have:
            \[
                u_{n+1} = 2u_n + (n+1) = 2\left( 2^{n+1} - (n+2) \right) + n + 1 = 2^{n+2} - (n+3),
            \]
            which shows the property holds for $n+1$.
        \end{minipage}
    }
    \caption{An example of an informal proof that needs to be formalized.}
    \label{fig:informal-proof-ex}
\end{figure}

Specifically, Figure~\ref{fig:informal-proof-ex} gives a motivating example from the miniF2F benchmark~\cite{zheng2022minif2f}.
The problem and its human-written informal proof are given, and it demonstrates a concise and correct inductive proof.
However, some intermediate steps, particularly the algebraic manipulations, are omitted in the informal proof.
Although the equation can be easily verified by human experts, it is challenging for LLMs to infer the missing details and generate a correct formal proof due to their limited symbolic reasoning capabilities.
The following Isabelle/HOL proof snippet illustrates an attempt by ChatGPT-3.5-turbo to formalize the given informal proof.
While the model correctly identifies the required transformations, it fails during the simplification of the term $2 * (n+2) - (n+1)$, mistakenly reducing it to $2n + 3$ instead of $n+3$.
This example highlights the limitations of LLMs in handling detailed algebraic reasoning without explicit guidance.

\begin{codeframe}
\begin{lstlisting}[xleftmargin=1em, language=Isar]
lemma induction_seq_mul2pnp1_forn:
  fixes n :: nat and u :: "nat \<Rightarrow> nat"
  assumes h0 : "u 0 = 0"
    and h1 : "\<And>(n::nat). u (n+1) = 2 * u n + (n+1)"
    and h2 : "u n = 2 ^ (n+1) - (n+2)"
  shows "u (n+1) = 2 ^ (n+2) - (n+3)"
proof -
  have "u (n+1) = 2 * u n + (n+1)" using h1 by simp
  also have "... = 2 * (2 ^ (n+1) - (n+2)) + (n+1)" using h2 by simp
  also have "... = 2 * 2 ^ (n+1) - (2 * (n+2)) + (n+1)"
    by (simp add: right_diff_distrib')
  also have "... = 2 ^ (n+2) - (2 * (n+2) - (n+1))"
    sledgehammer
  (* an error in simplification *)
  also have (*@\error{"... = 2 \^{} (n+2) - (2*n + 3)"}@*) by auto
  also have "... = 2 ^ (n+2) - (n+3)" by auto
  finally show ?thesis by blast
qed
\end{lstlisting}
\end{codeframe}



To address this, we propose the use of autoformalization agents equipped with enhanced capabilities for symbolic reasoning.
More specifically, autoformalization agents break down the process into manageable steps: (i) generating proof outlines, (ii) filling intermediate steps using external tools, and (iii) integrating and refining proofs.
To elaborate, the agent first constructs a high-level proof outline, capturing the main steps of the informal proof, while leaving placeholders for missing intermediate steps.
This outline aligns with the informal proof structure and serves as a blueprint for the following formalization process.
For the missing details, especially those involving symbolic reasoning or algebraic manipulations, the agent delegates the task to external tools like computer algebra systems (CAS).
These tools can perform accurate transformations on the mathematical expressions, ensuring the correctness of the derived intermediate steps.
Once the intermediate steps are derived, the agent integrates them into the proof outline, filling in the placeholders and completing the formalization.
In case the agent still encounters gaps in certain steps, it iteratively refines the proof by revisiting the informal proof and consulting external tools.
In this way, the autoformalization agents can leverage the strong symbolic reasoning capabilities of external tools to fill in the missing details in the informal proofs, thus bridging the gap between informal and formal proofs and specifications.





\subsection{LLM for Theorem Proving}


Among all formal analysis techniques, theorem proving stands out for its capability to handle complex state spaces, abstract specifications, and highly intricate systems. Unlike model checking (we will discuss subsequently) which is primarily designed for finite models and faces challenges with state space explosion, theorem proving excels in leveraging mathematical reasoning to establish properties that hold universally.
This capability has been successfully demonstrated in critical systems, such as CompCert~\cite{leroy2009formal}, a formally verified C compiler that guarantees the correctness of compiled code, and seL4~\cite{klein2009sel4}, a microkernel with rigorous proofs of memory safety, functional correctness, and security properties.
These examples showcase the unmatched flexibility theorem proving offers in system verification, making it a powerful tool for ensuring system correctness. In this section, we explore how LLMs can augment each sub-task within the theorem-proving framework, improving efficiency and enhancing accessibility for novice users throughout the entire proving process.


\subsubsection{Premise Selection}

Retrieving relevant facts from a large collection of lemmas is a critical task in theorem proving, and this process is known as premise selection.
This task is typically done manually by explicitly specifying the used lemmas in the proof scripts, which often requires trial and error and deep domain knowledge, making it time-consuming and error-prone.
Some powerful automation techniques in interactive theorem provers (ITPs) also need premise selection to first filter out irrelevant lemmas from the large search space.
For example, Sledgehammer~\cite{bohme2010sledgehammer}, an effective tool for Isabelle/HOL~\cite{paulson1994isabelle}, collects relevant facts from the background theories and sends them to external automatic theorem provers (ATPs) and SMT solvers to find proofs.
This process involves premise selection to identify the most relevant lemmas that can help in proving the current goal. For example, Sledgehammer usually selects about 1,000 lemmas out of tens of thousands of premises.
Some heuristics~\cite{megn2009lightweight} and machine learning techniques~\cite{kuhlwein2013mash} like naive Bayes are used in Sledgehammer for relevant fact selection.
A recent work~\cite{mikula2024magnushammer} proposed using transformer models to learn the relevance of lemmas for premise selection, which improved the success rate of Sledgehammer by 13\% on the miniF2F benchmark~\cite{zheng2022minif2f}.

Our insight is that LLM agents can further improve the premise selection process by leveraging their code understanding capabilities.
Premise selection is fundamentally different from other tasks like code retrieval, where cosine similarity is often used to identify candidates with high syntactic resemblance.
While such approaches are effective in retrieval contexts, they fall short in theorem proving.
Lemmas with similar syntactic structures to the proof goal may not necessarily be the most useful, and conversely, the most helpful facts may exhibit little syntactic similarity.
This highlights the need for a retrieval approach grounded in the semantic understanding of lemmas, rather than relying solely on syntactic features as most current methods do.

LLMs, with their strong code comprehension capabilities, offer a way to address this gap.
They can infer the meaning of a lemma from its name, its definition, and its contextual information, mimicking the reasoning process of a human expert.
For instance, a human expert can intuitively assess whether a given lemma is likely to be helpful for a particular proof goal.
However, the sheer number of lemmas in large proof libraries makes it impossible for experts to manually evaluate and rank all possible candidates.
By contrast, LLM agents can efficiently scale this process.
We can first collect definitions and contextual information of lemmas and ask LLM agents to generate semantic descriptions in natural language for each lemma, forming a knowledge base for premise selection.
Then, given a proof goal, LLM agents comprehend the goal and generate a semantic representation, which is used to query the knowledge base for relevant lemmas.
This approach effectively simulates the expert's intuition across a vast number of possibilities, fully leveraging semantics information to guide the premise selection process.

\subsubsection{Proof Generation}

Proof step generation is the central task in theorem proving, where the objective is to predict one or more proof steps to construct a valid proof for a given theorem.
Many pioneering works on LLM-based proof generation~\cite{polu2020generative,polu2022formal,han2021proof} approach this problem as a language modeling task and train LLMs on large-scale proof corpora to predict the next proof step.
Various techniques have been developed to improve the quality of generated proofs.
For instance, learning to invoke ATPs to discharge subgoals~\cite{jiang2022thor}, repairing failed proof steps by querying LLMs with the error message~\cite{first2023baldur}, and predicting auxiliary constructions to simplify proofs~\cite{trinh2024solving} have all demonstrated significant potential.

However, real-world verification scenarios present challenges that go beyond these methods.
Human experts, for instance, do not solely rely on immediate proof context or predefined strategies.
Instead, they first have a high-level proof plan in mind and frequently need to consult the definitions of important concepts or theorems during the proof process.
Additionally, experts often employ a trial-and-error approach, iteratively refining their methods to construct a valid proof.
This highlights a limitation of current LLMs when used as standalone tools: while they excel at producing plausible proof steps, they lack the broader strategic reasoning and adaptability.
This gap makes it difficult for LLMs to consistently surpass human performance in proof generation tasks.

To address these limitations, we propose a shift toward LLM agents that more closely emulate human experts in their proof strategies.
In contrast to standalone LLMs, these agents integrate multiple capabilities, allowing them to reason, adapt, and interact during the proof process.
This distinction can be articulated through the following two key features:

\smallskip
\noindent
\textbf{Feature 1. Explicit Proof Intentions.}
A defining feature of LLM agents is their ability to generate both proof steps and explicit proof intentions---statements that explain the reasoning or goals underlying each step.
This additional layer of information is critical for improving both automated and human-driven refinement.
When a proof step fails, the agent can use the intention, along with error feedback, to attempt a proof repair.
Even if the repair is unsuccessful, the intention provides valuable insights for human users, streamlining their efforts to identify and resolve the issue.

\noindent
\textbf{Feature 2. Dynamic Retrieval of Relevant Knowledge.}
LLM agents go beyond the immediate context by incorporating mechanisms to retrieve definitions, lemmas, or related theorems from knowledge bases.
This mimics how human experts consult reference materials during the proof process but with significantly greater efficiency and scale.
By dynamically identifying and incorporating relevant information, the agent can address gaps in its internal knowledge, enabling it to construct proofs that require broad or specialized domain understanding.

\smallskip
Many existing methods, such as delegating specific tasks to automated tools like Sledgehammer, and introducing auxiliary lemmas, can be seen as attempts to replicate human expert strategies.
LLM agents can unify these capabilities into a framework, allowing them to reason about when and how to apply these techniques.
By combining these methods with explicit intentions and dynamic knowledge retrieval, LLM agents can achieve a level of adaptability and strategic reasoning that is closer to human expertise.

\subsubsection{Case Study on LLMs with Coq}

To illustrate our perspective, we illustrate our recent exploration of the interaction between LLMs and Coq. 
Coq~\cite{huet1997coq} is a classic proof assistant based on constructive type theory, supporting functional programming and formal specification. 
The integration of Coq with an LLM agent involves several key steps:

\smallskip
\noindent
{\bf Step 1. Natural Language Understanding.} The LLM agent receives natural language input from the user, typically in the form of a mathematical theorem, conjecture, or problem.

\noindent
{\bf Step 2. Formalization of the Problem.} The LLM agent translates the natural language problem into Coq’s formal language. This includes defining types, propositions, and functions necessary for the formulation of the theorem.

\noindent
\textbf{Step 3. Proof Construction.} The LLM agent collaborates with Coq to construct proofs, utilizing Coq’s interactive features to propose proof steps that are subsequently verified or refined.


\noindent
\textbf{Step 4. Proof Verification and Feedback.} Once the proof is constructed, Coq verifies its correctness. The LLM agent translates the verified proof back into natural language, providing a comprehensible explanation of the result to the user.


\paragraph{Example}
Consider the following natural language query posed to the LLM agent: 
\begin{quote}
    \textit{``Prove that the sum of two even numbers is even.''}
\end{quote}

\smallskip
The LLM agent first translates this into a formal Coq language:

\begin{codeframe}
\begin{lstlisting}[xleftmargin=1em, language=Coq]
Theorem sum_of_even_numbers : forall n m : nat,
  even n -> even m -> even (n + m).
\end{lstlisting}
\end{codeframe}

The user needs to check the correctness of the coq code to guarantee consistency between the natural language and coq code.
Then, the LLM agent will search the related theorems and lemmas, and construct the proof as follows:

\begin{codeframe}
\begin{lstlisting}[xleftmargin=1em, language=Coq]
Proof.
  intros n m Hn Hm.
  unfold even in *.
  destruct Hn as [k Hk].
  destruct Hm as [l Hl].
  exists (k + l).
  rewrite Hk, Hl.
  ring.
Qed.
\end{lstlisting}
\end{codeframe}
Coq verifies this proof, and the LLM agent explains and responds in natural language:
\begin{quote}
\textit{``The proof shows that if two numbers are even, their sum is also even by using basic properties of addition and the definition of even numbers.''}
\end{quote}


\subsection{LLM for Model Checking} 

Model checking is a formal verification technique that systematically explores a system's state space to check if the system satisfies specified properties, such as safety and liveness. It is particularly effective for finite-state systems, providing automated detection of logical errors like deadlocks or critical system property violations. However, traditional model checking faces great limitations, including scalability challenges for large systems and the complexity involved in system modeling and property formalization.

In this section, we illustrate how model checking agents can alleviate above limitations by combining the strengths of LLMs. Leveraging the respective advantages, a model checking agent accepts a system description in natural language from the user, generates corresponding processes using an LLM, and refines them iteratively based on feedback from the model checker. We note that this refinement can be facilitated by an Automated Prompt Engineer (APE)~\cite{cheng2023black,zhou2022large}, which optimizes the interaction between the LLM and the model checker. This synergy not only streamlines the model checking process but also enhances its accessibility for users without extensive expertise in formal verification.




\subsubsection{Overview}
The overall workflow for building a model checking agent should include the following steps: 

\smallskip
\noindent
\textbf{Step 1. Natural Language Input.} Users provide a description of a system in natural language, specifying desired behaviors or properties. For instance, they might describe a mutual exclusion protocol or outline expected system behaviors.

\noindent
\textbf{Step 2. LLM Planning.} An LLM with strong reasoning capabilities translates the natural language input into a detailed, structured plan. This plan breaks down the system description into specific instructions, including a one-to-one mapping of logical steps, such as defining variables, specifying state transitions, and drafting preliminary assertions.

\noindent
\textbf{Step 3. LLM Code (Assertion) Completion.}
Following the structured plan, a specialized LLM trained in syntax and logic generates the code and assertions required to implement the system. This generation is treated as a completion task, where the model fills in the details based on the planning phase rather than creating the code from scratch. The structured approach enhances precision and reduces the likelihood of errors in the generated outputs.

\noindent
\textbf{Step 4. Verification.}
The generated code and assertions are submitted to an automated verification tool. This tool evaluates the artifacts to ensure they comply with the logical properties and behaviors specified in the initial input. 

\noindent
\textbf{Step 5. Feedback Generation.}
Feedback from the formal verification tool highlights issues such as syntax errors or logical inconsistencies. This feedback is processed by an APE, which refines the LLM's prompts and provides updated instructions.

\noindent
\textbf{Step 6. Iterative Improvement.}
The LLM iteratively refines the code and assertions based on feedback until all formal verification checks are successfully passed. Through this iterative process, the framework ensures that the final system model is logically sound, adheres to the desired specifications, and is ready for deployment.

\smallskip
In the following part, we demonstrate our model checking agent framework utilizing the widely adopted model checker, Process Analysis Toolkit (PAT)~\cite{sun2008model,LiuSD11PAT3}. 
This demonstration highlights how the integration of formal methods and LLMs can be interactively harnessed to develop the PAT Agent, showcasing the potential of this synergy in enhancing verification processes. 


\subsubsection{PAT Agent}
Process Analysis Toolkit (PAT) is a formal verification tool designed to model, simulate, and verify concurrent and real-time systems. It supports the verification of key properties such as deadlock-freeness, reachability, and refinement, addressing critical correctness and reliability concerns in system design. With applications spanning domains such as vehicle and aircraft safety, resource optimization, and complex system analysis, PAT provides a robust foundation for developing a model checking agent.

In the following motivating example, we illustrate how the PAT Agent can be used to derive verified system constructs, starting from natural language instructions.

\paragraph{Example}
In the development of a car system, ensuring that the key is never locked inside the car is a critical requirement. This safeguard is essential for user convenience and safety, as locking the key inside could result in costly locksmith services, delays, or emergencies in dangerous areas. The system must ensure logical consistency in different operations to avoid such consequences.


\smallskip
We begin by asking an LLM (gpt-4o-2024-08-06) to directly generate a formal model based on a high-level description. While the LLM has learned syntax rules and demonstrates effective planning capabilities, the generated model contains a critical logic flaw: it allows the key to be locked inside the car. This issue arises due to  hallucination by GPT-4o, which incorrectly assumes that placing the key into a currently locked car is a valid scenario, contradicting commonsense reasoning. 

\begin{codeframe}
\begin{lstlisting}[xleftmargin=1em, basicstyle=\ttfamily\footnotesize, showlines=false]
[key == i && owner[i] == near]leavekey{key = incar;}
\end{lstlisting}
\end{codeframe}

Consequently, the resulting system design deviates from the intended behavior, compromising its reliability and safety.


To address this issue, we use PAT to formally verify the generated system. PAT identifies an error trace (i.e., a sequence of operations) that leads to the key being locked inside the car, providing feedback that highlights flaws in the logic governing key and door operations. By analyzing this error trace, the LLM identifies the aforementioned logical flaw and corrects it by imposing stricter restrictions on when the key can be left inside the car. Additionally, it establishes clear conditions for locking the door, ensuring alignment with the intended system behavior.

\begin{codeframe}
\begin{lstlisting}[xleftmargin=1em, basicstyle=\ttfamily\footnotesize, showlines=false]
[key == i && owner[i] == near && door == open]
leavekey{key = incar;}
\end{lstlisting}
\end{codeframe}

\begin{codeframe}
\begin{lstlisting}[xleftmargin=1em, basicstyle=\ttfamily\footnotesize, showlines=false]
[(owner[i] == near && key == i) || owner[i] == in]
lockdoor.i{door = lock;}
\end{lstlisting}
\end{codeframe}

The refined model, guided by PAT's feedback, results in a system that can now be formally verified to ensure the key will never be locked inside the car. This example highlights the powerful synergy between formal verification and LLM-driven development. LLMs ease the process of system development, while tools like PAT provide rigorous guarantees by identifying and rectifying logical inconsistencies. Together, they create a robust and user-friendly approach to formal system development, ensuring critical requirements are met with precision.

\begin{figure*}[ht!]
 \begin{centering}
  \includegraphics[width=0.7\linewidth]{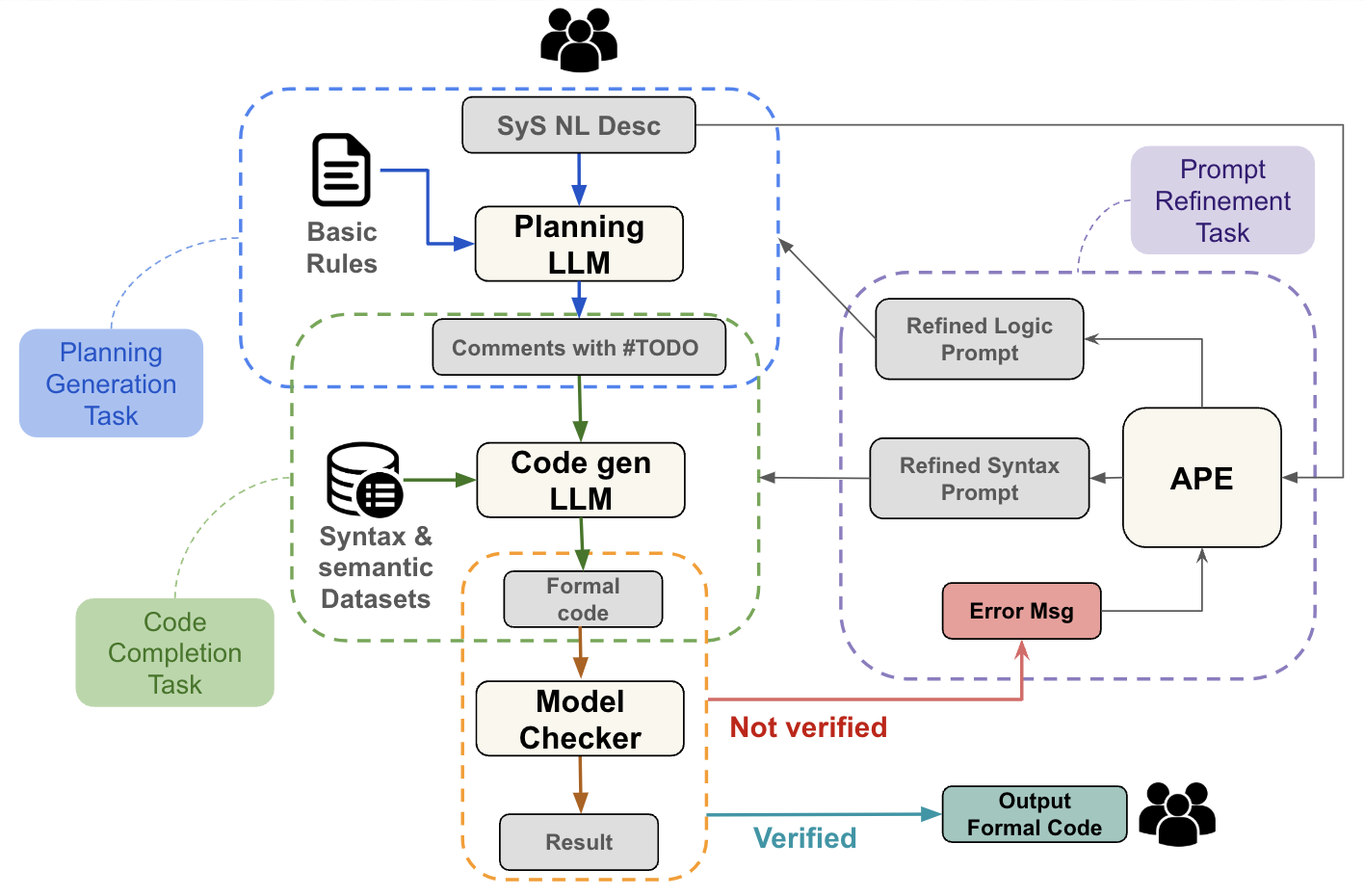}
  \par\end{centering}
 \caption{PAT Agent Prototype.}
 \label{fig:prototype}
\end{figure*}

\paragraph{Prototype} To implement the PAT model checking agent, we build it on the LangChain pipeline, leveraging its streamlined implementation and modularized memory system to efficiently manage intermediate states and iterative refinements during formal verification. This allows seamless integration of LLM capabilities with formal verification tools. We illustrate the prototype in Figure~\ref{fig:prototype}.

The PAT Agent prototype follows a structured workflow that seamlessly integrates formal verification with LLM capabilities. Starting with natural language descriptions provided by the user, the Planning LLM translates these into structured instructions, specifying logical components such as state transitions and properties to verify. These instructions are passed to the Code Generation LLM, which generates formal code with knowledge learned from the syntax and semantic datasets as a completion task. The generated code is then verified with the PAT model checker, evaluating key properties such as deadlock-freeness and reachability. If verification fails, the APE refines the instructions based on feedback from PAT, iteratively improving the code until it meets the specified properties. By uniting the strengths of LLMs and formal verification, the PAT Agent exemplifies how model checking agents can enable reliable and accessible system development.

\subsubsection{Other Model Checking Agents.} We acknowledge that such an agent framework is highly generalizable and can be adapted for tools like Alloy Analyzer~\cite{jackson2000automating}, PRISM~\cite{kwiatkowska2002prism}, UPPAAL~\cite{behrmann2004tutorial}, etc. Its modular design enables the Planning LLM to learn tool-specific semantics and the Code Generation LLM to produce corresponding formal code and assertions. By tailoring feedback and refinement loops for each tool, the framework seamlessly integrates LLM-driven development with diverse formal verification processes.


\section{Unifying the Power of FMs and LLMs}\label{sec:fm_llm}

This section highlights the synergy between FMs and LLMs, combining the rigor of formal methods with the flexibility and intelligence of LLMs. 
By integrating these two approaches, we can create robust systems that enhance both the trustworthiness and efficiency of software engineering processes while adapting to complex and real-world problems. We demonstrate this idea through an application: trustworthy code generation.

We acknowledge that using an LLM for program synthesis involves leveraging its ability to understand and generate code snippets, solve programming problems, and create software systems based on high-level requirements. 
This integration aims to leverage the LLM agent’s ability to understand natural language, translate program specifications and code into formal properties, and assist in verifying correctness using automated verification tools.

\subsection{Challenges of LLMs for Program Synthesis}
LLMs have made rapid advancements in mathematics, reasoning, and programming~\cite{zhao2023survey,romera2023mathematical}, significantly reshaping the landscape of software development and computational problem-solving. Industrial-grade LLMs like GPT-4~\cite{openai2023gpt4} and tools like GitHub Copilot~\cite{copilot} have emerged as powerful assistants for programmers, automating coding-related tasks and achieving performances exceeding the 50th percentile in competitive programming contests~\cite{pandey2024transformingsoftwaredevelopmentevaluating}. These achievements highlight the transformative potential of LLMs in enhancing productivity and reducing manual effort in coding.

However, despite these advancements, LLMs face significant challenges, notably the issue of hallucination.
This phenomenon undermines their reliability in critical contexts. User studies~\cite{vaithilingam2022expectation,ding2023static} have revealed that programmers often struggle to trust and debug LLM-generated code due to the opaqueness and lack of control over the generation process. Alarmingly, empirical research indicates that over half of ChatGPT's responses to programming-related queries contain inaccuracies~\cite{Kabir24}, exacerbating concerns about their utility in professional and high-stakes environments. Even worse, recently, mathematicians have proven that hallucinations in LLMs are unavoidable~\cite{xu2024}.

These challenges underscore the urgent need for strategies to mitigate hallucination and improve the transparency and trustworthiness of LLM-generated code, paving the way for their broader adoption in critical programming workflows.

\subsection{Program Refinement}
The task of program refinement has a rich history, dating back to foundational works such as  ~\cite{dijkstra1976discipline,floyd1993assigning,hoare1969axiomatic}.
The related theories ~\cite{ProgramFromSpecification, back1990refinement} are based on Hoare logic and the calculus of weakest preconditions. 
Recent advancements have extended these foundational ideas by formalizing the refinement calculus within interactive theorem provers. Notable examples include the work in Isabelle~\cite{foster2020differential} and Coq~\cite{Refinecoq, MechanizedTheoryPR}, leveraging the expressive power of these tools to mechanize and verify program refinements. The Coq proof assistant, in particular, has been instrumental in encoding and verifying complex refinement relations~\cite{barras1999coq}.

Formally, the refinement relation between programs is grounded in their weakest preconditions, enabling precise definitions and automated verification~\cite{Butler}. 
This formalism ensures correctness and supports incremental development by iteratively refining abstract specifications into concrete implementations while maintaining rigorous correctness guarantees. 
These developments mark a significant step forward in the use of formal methods for program synthesis and verification.
For program $S$ and postcondition $P$, $wp(S,P)$ represents the weakest precondition where S is guaranteed to terminate in a state satisfying $P$.
Program $S_0$ is refined by $S_1$ denoted as $S_0 \sqsubseteq S_1$, iff
\begin{equation}
    \forall P, wp(S_0,P) \to wp(S_1,P)
\end{equation}
which states that $S_1$ will preserve the total correctness of program $S_0$.
The program refinement can be established in a linear sequence:
\begin{equation}
    S_0 \sqsubseteq S_1 \sqsubseteq S_2 \sqsubseteq S_3 ... \sqsubseteq S_n
\end{equation}
which shows the refinement $S_0 \sqsubseteq S_n$ with the transitivity of the refinement relation.

\subsection{Integrating Formal Program Refinement with LLMs}

The refinement calculus ~\cite{ProgramFromSpecification,back1990refinement,carrington1998program, MechanizedTheoryPR,swierstra2016proposition} provides a rigorous framework for the stepwise refinement method of program construction. This approach involves specifying a program's required behavior as an abstract, non-executable specification, which is systematically transformed into an executable program through a series of correctness-preserving steps. By ensuring that each step maintains the program's intended semantics, the refinement calculus offers a sound methodology for constructing reliable software.

\subsubsection{Technical Challenges}
Despite its strengths, the refinement process is traditionally performed manually, making it a labor-intensive and error-prone task. The reliance on human effort to translate abstract specifications into executable code not only slows down development but also increases the risk of introducing subtle mistakes during transformation. This bottleneck has long been a barrier to scaling the benefits of refinement calculus to larger, more complex software systems.

Given the advancements in LLMs, integrating their powerful code-generation capabilities into the refinement process presents an exciting opportunity. LLMs can assist in automating parts of the transformation, bridging the gap between high-level specifications and executable implementations while adhering to the principles of correctness-preserving refinement. This integration could significantly reduce the manual effort required, streamline the refinement process, and open new avenues for leveraging LLMs in formal methods and program synthesis.

\begin{figure*}[t]
\centering
\includegraphics[width=0.7\textwidth]{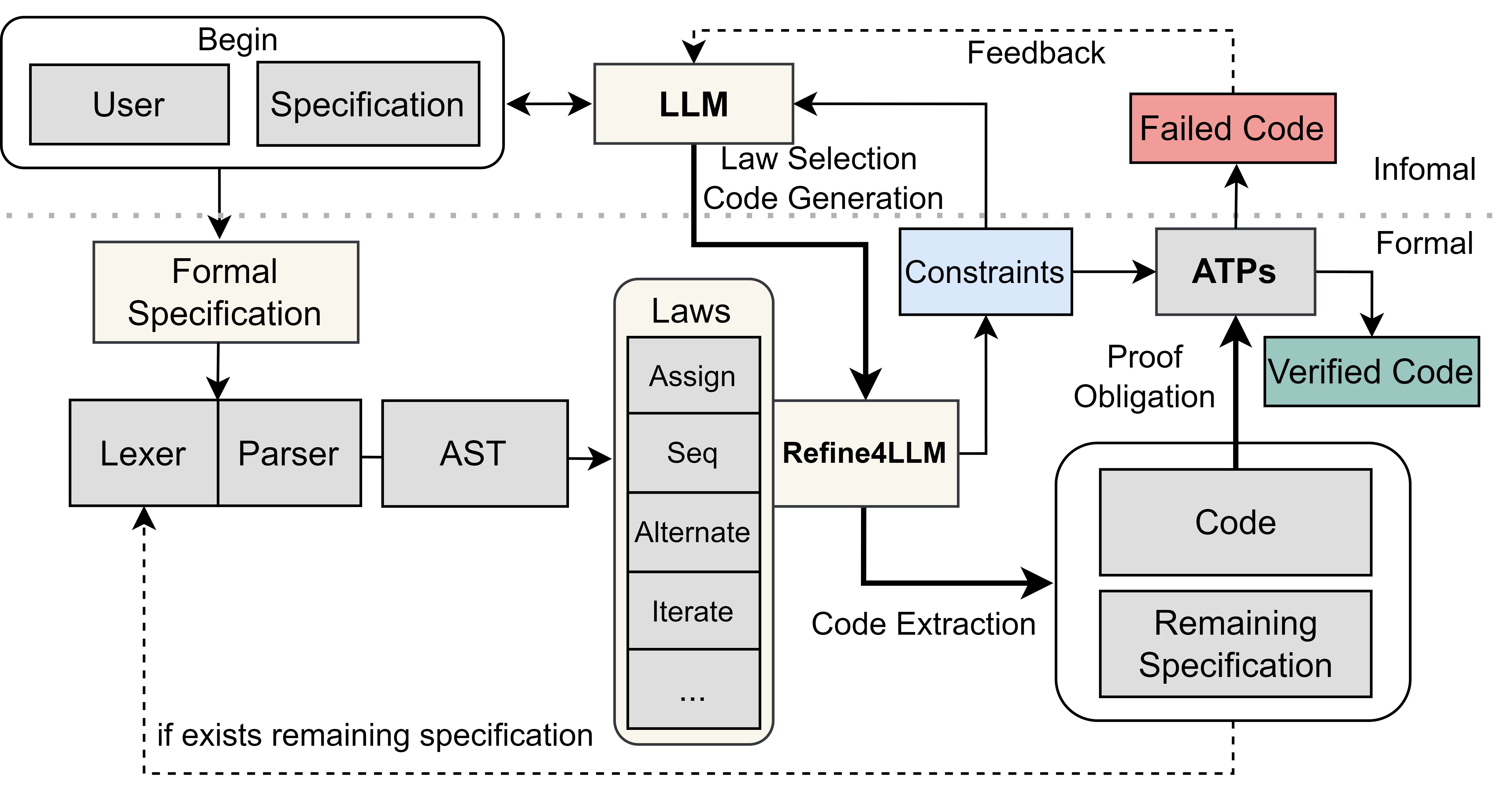}
\caption{Overview of the tool that combines LLMs and program refinement.}
\label{fig:approach}
\Description{}
\end{figure*}

\subsubsection{Our Insights.}
\autoref{fig:approach} shows an overview of our approach. 
Generally, it can be divided into two parts: the formal and informal systems.
We define the formal specification language and program language in the formal system.
The formal specification will first be transformed into an abstract syntax tree. 
Then, with one specific law, it will formally transform the specification into the new specification and build the proviso constraints of the law.
Finally, after extracting the generated code from LLM, it will automatically build the proof obligations and the verification scripts for ATPs to verify.
ATPs will try to verify the scripts automatically and output the success or error message.
If the code fails to verify, it will provide feedback to the LLM with possible counterexamples.

In the informal system, the user needs to first formalize the natural language specification into a formal specification with the aid of the LLM.
Secondly, it will build the prompt and describe the refinement laws to the LLM.
Then, the LLM selects one refinement law based on the specification's description and constraints and generates the associate code.
The LLM will regenerate the code based on the verification result.
If multiple failures occur, 
our method will backtrack
to the last refinement step, interact with the LLM again to choose another refinement law, re-generate the associated code, and then proceed as previously.

\section{Conclusion and Discussion}\label{sec:con}
This paper presents a roadmap for the fusion of Large Language Models and Formal Methods to construct next-generation trustworthy AI agents/systems. By leveraging the complementary strengths of LLMs (adaptability, natural language processing, and intelligence) and FMs (rigorous reasoning, verification, and provable guarantees), we propose a synergistic framework to address the limitations of both paradigms. Through case studies and conceptual explorations, we demonstrate the transformative potential of integrating these technologies in applications such as program synthesis. This convergence not only paves the way for robust AI systems capable of solving complex real-world problems but also provides a foundational step toward bridging neural learning and symbolic reasoning for enhanced trustworthiness in AI-driven systems.

Indeed, the tight integration of LLMs and FMs raises several significant technical and practical challenges. On the technical front, achieving seamless communication between these paradigms demands a sophisticated mechanism for translating natural language based specifications into formal representations and vice versa. This translation must be accurate, scalable, and contextually adaptable to maintain the integrity of both paradigms. Additionally, LLMs' inherent issues with hallucinations and logical inconsistencies pose a barrier to their deployment in safety-critical domains. Incorporating certification mechanisms is essential for mitigating these risks and ensuring system reliability. 
From a practical perspective, the usability and accessibility of these integrated systems require careful consideration. 
Specifically, the adaptability must be carefully balanced with robust safeguards to address potential misuse or misinterpretation of outputs, particularly in high-stakes scenarios. An equally critical concern is the protection of client privacy, as formal verification tools and LLM agents often process sensitive or proprietary data. Ensuring privacy requires implementing secure data-handling protocols, such as local deployment of tools, encrypted communication, and adherence to strict data governance policies. Future research should focus on developing user-friendly tools that not only retain the rigor of formal methods and simplify interactions for users without formal engineering expertise but also incorporate robust privacy-preserving mechanisms to safeguard client information.

\bibliographystyle{ACM-Reference-Format}

\bibliography{ref}



\end{document}